\documentclass[12pt]{article}

\usepackage{color}

\usepackage{amsfonts,amstext,amsmath,amssymb,amsthm}
\usepackage{multirow}
\usepackage{subfigure}
\usepackage[utf8]{inputenc} 
\usepackage[T1]{fontenc}    
\usepackage{hyperref}       
\usepackage{url}            
\usepackage{booktabs}       
\usepackage{amsfonts}       
\usepackage{nicefrac}       
\usepackage{microtype}      
\usepackage{lipsum}
\usepackage{graphicx}
\usepackage{wrapfig}
\graphicspath{ {./images/} }
\usepackage{natbib}
\usepackage{algorithm,algorithmic}
\usepackage[symbol]{footmisc}
\usepackage{setspace}   
\usepackage{geometry}
\usepackage[dvipsnames]{xcolor}

\def\T{{ \mathrm{\scriptscriptstyle T} }}
\newcommand{\norm}[1]{\|#1\|}

\newcommand{\RR}{\mathbb{R}}

\newtheoremstyle{mytheoremstyle} 
    {\topsep}                    
    {\topsep}                    
    {\normalfont}                   
    {}                           
    {\bfseries}                   
    {.}                          
    {.5em}                       
    {}  

\theoremstyle{mytheoremstyle}

\ifx\BlackBox\undefined
\newcommand{\BlackBox}{\rule{1.5ex}{1.5ex}}  
\fi

\ifx\QED\undefined
\def\QED{~\rule[-1pt]{5pt}{5pt}\par\medskip}
\fi

\ifx\proof\undefined
\newenvironment{proof}{\par\noindent{\bf Proof\ }}{\hfill\BlackBox\\[2mm]}
\fi

\ifx\theorem\undefined
\newtheorem{theorem}{Theorem}
\fi
\ifx\example\undefined
\newtheorem{example}{Example}
\fi
\ifx\property\undefined
\newtheorem{property}{Property}
\fi
\ifx\lemma\undefined
\newtheorem{lemma}{Lemma}
\fi
\ifx\proposition\undefined
\newtheorem{proposition}{Proposition}
\fi
\ifx\remark\undefined
\newtheorem{remark}{Remark}
\fi
\ifx\corollary\undefined
\newtheorem{corollary}{Corollary}
\fi
\ifx\definition\undefined
\newtheorem{definition}{Definition}
\fi
\ifx\conjecture\undefined
\newtheorem{conjecture}{Conjecture}
\fi
\ifx\fact\undefined
\newtheorem{fact}{Fact}
\fi
\ifx\claim\undefined
\newtheorem{claim}{Claim}
\fi
\ifx\assumption\undefined
\newtheorem{assumption}{Assumption}
\fi
\ifx\condition\undefined
\newtheorem{condition}{Condition}
\fi

\begin{document}

\title{Causal Structural Learning from Time Series: \\ A Convex Optimization Approach} 

\author{
 Song Wei, \quad  Yao Xie\\
  H. Milton Stewart School of Industrial and Systems Engineering\\
Georgia Institute of Technology, Atlanta, Georgia, 30332, U.S.A. \\
}

\date{\vspace{-20pt}}

\maketitle

\footnotetext{This work is partially supported by an NSF CAREER CCF-1650913, and NSF DMS-2134037, CMMI-2015787, DMS-1938106, and DMS-1830210. Correspondence to:  Yao Xie <yao.xie@isye.gatech.edu>.}

\begin{abstract}
Structural learning, which aims to learn directed acyclic graphs (DAGs) from observational data, is foundational to causal reasoning and scientific discovery. Recent advancements formulate structural learning into a continuous optimization problem; however, DAG learning remains a highly non-convex problem, and there has not been much work on leveraging well-developed convex optimization techniques for causal structural learning. We fill this gap by proposing a data-adaptive linear approach for causal structural learning from time series data, which can be conveniently cast into a convex optimization problem using a recently developed monotone operator variational inequality (VI) formulation. Furthermore, we establish non-asymptotic recovery guarantee of the VI-based approach and show the superior performance of our proposed method on structure recovery over existing methods via extensive numerical experiments.
\end{abstract}

\section{Introduction}\label{sec:intro}

Causal discovery, which aims to capture the interactions among events of interest using directed acyclic graphs (or Bayesian networks), is a crucial part of scientific discovery \citep{pearl2009causality} and has drawn much attention recently. With advanced data acquisition techniques, we usually observe time series data in many modern applications, posing both opportunities to learn a dynamic Bayesian network and challenges in finding an efficient approach for learning a directed acyclic graph (DAG) from serially correlated data \citep{pamfil2020dynotears}.

However, learning DAGs from observational data, i.e., the structural learning problem, is NP-hard due to the combinatorial acyclicity constraint \citep{chickering2004large}, motivating many research efforts in finding efficient approaches for learning DAGs.
Recently, \citet{zheng2018dags} proposed a continuous differentiable characterization of DAG, which formulates the DAG learning problem into a constrained continuous optimization problem; they applied augmented Lagrangian method to transfer constraint into penalty and achieved efficient DAG learning.
Later on, \citet{ng2020role} proposed to treat the non-convex DAG characterization as penalty and proved asymptotic recovery guarantee for linear Gaussian models.

On the other hand, recently much work has been done on causal discovery from time series; notable contributions include 
Fourier-transform based time series approach for continuous-time Hawkes process models \citep{etesami2016learning}. 
However, existing works have been mostly focusing on Granger causality, which has been deemed less useful due to the lack of DAG structure in the estimated causal graph. To fix this issue, \citet{pamfil2020dynotears} 
leveraged the continuous DAG characterization as the constraint in structural vector autoregressive models for Granger causal discovery and solved the constrained optimization problem via augmented Lagrangian method as \citet{zheng2018dags} did. 
Despite those recent advancements, DAG learning remains a non-convex problem. Thus, how to leverage the well-developed convex optimization techniques to learn a DAG largely remains an open problem.

\begin{figure}[!htp]
\centering
{\includegraphics[width = .75\textwidth]{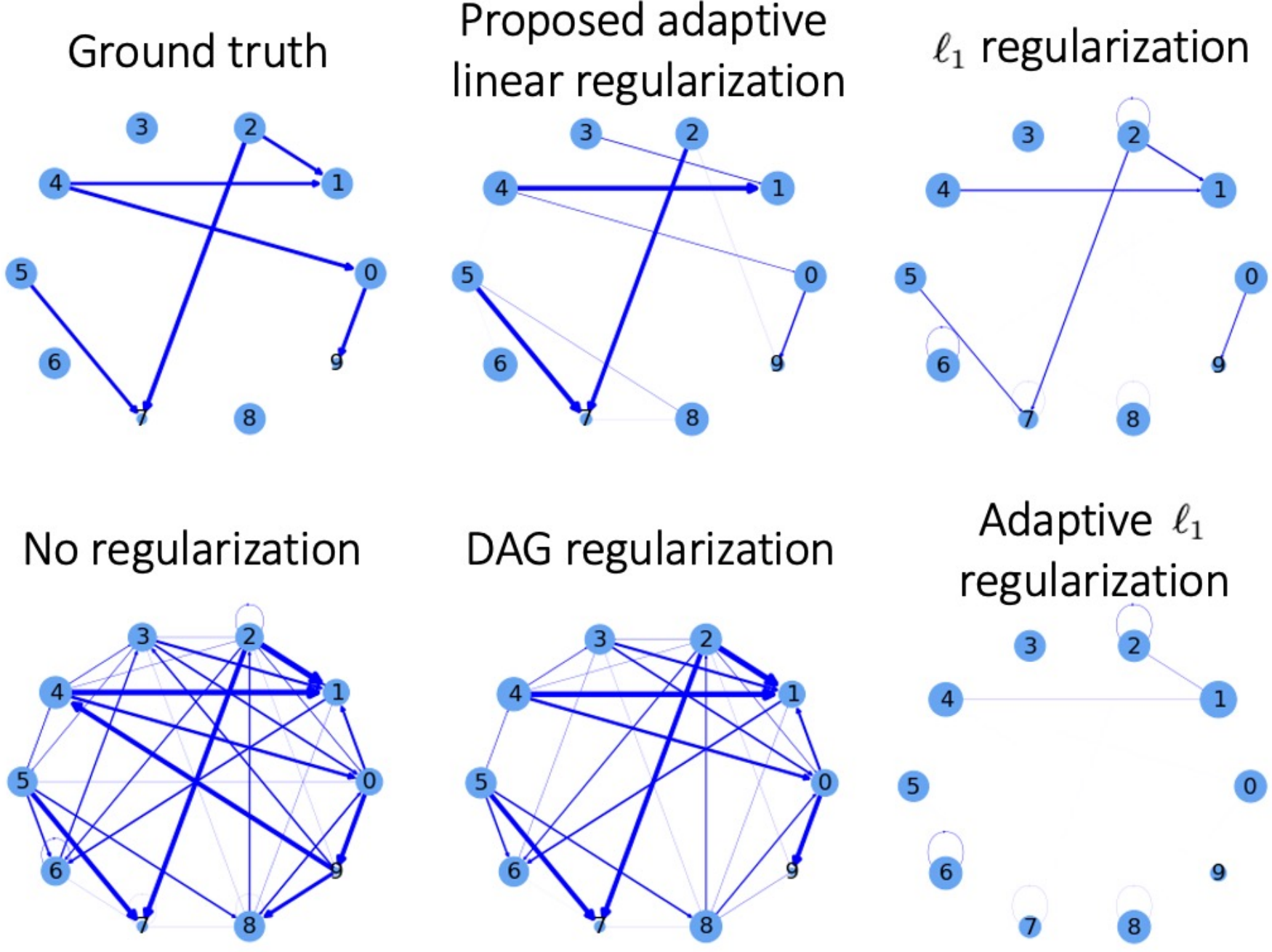}}
\caption{Visualization of estimated graphs, where the size of the node is proportional to the background intensity and the width of the edge is proportional to the exciting coefficient magnitude. We consider a graph with $d_1 = 10$ nodes and time horizon $T = 500$; the data is generated via our GLM with exponential link. We compare various types of regularization (specified on top of each panel). The Structural Hamming Distances between the estimated graph and the ground truth are 39 (no regularization), {\it 7 (proposed)}, 21 (DAG regularization \citep{zheng2018dags,ng2020role}), 25 ($\ell_1$ regularization) and 12 (adaptive $\ell_1$ regularization \citep{zou2006adaptive}), respectively. Our proposed data-adaptive linear regularization achieves the best graph structure recovery.}\label{fig:exp1_fitted_graph_illus}
\end{figure}

In this work, we present a generalized linear model (GLM) based approach for causal discovery from time series data, while seeking the DAG structure via a novel data-adaptive linear regularizer. 
Furthermore, we cast the DAG structural learning problem into a convex optimization program by a monotone operator variational inequality (VI) formulation. 
The convex formulation enables us to establish non-asymptotic performance guarantee for a wide range of non-linear link functions via recent advances in VI-based signal recovery \citep{juditsky2019signal,juditsky2020convex}. 
We provide extensive numerical experiments to show the competitive performance of the proposed method and observe that our approach achieves more performance gain in the presence of limited data (see Figure~\ref{fig:exp1_fitted_graph_illus} for illustration).

\paragraph{\it Literature.}
Efficient structural learning of a DAG is the heart of scientific discovery in many fields, e.g., biology \citep{sachs2005causal},
genetics \citep{zhang2013integrated}, and so on. In particular, in causal reasoning, structural causal model based causal discovery methods oftentimes boil down to maximizing a score function within the DAG family \citep{glymour2019review}. 
There is rich literature in DAG learning: 
\citet{yuan2019constrained} proposed to use indicator function to enumerate and eliminate all possible directed cycles; to efficiently solve such problem, they used truncate $\ell_1$-function as a continuous surrogate of indicator function and proposed to use alternating direction method of multipliers to numerically solve it.
\citet{manzour2021integer} transferred indicators into binary variables and leveraged mixed integer programming to solve it.
There are also dynamic programming based approaches, e.g., \citet{loh2014high}, but they are not scalable in high dimensions unless coupled sparse structure, e.g., $A*$ Lasso \citep{xiang2013Lasso}. 
Another line of research follows the continuous DAG characterization by \citet{zheng2018dags}; in addition to aforementioned developments, notable extensions along this direction includes a discrete backpropagation method, exploration of low-rank structure \citep{fang2020low} and neural DAG learning \citep{yu2019dag,ke2019learning,lachapelle2019gradient}. We refer readers to \citet{scanagatta2019survey,kitson2021survey,vowels2022d} for systematic surveys on structural learning and causal discovery.

\section{Background}\label{sec:background}

\subsection{Problem set-up}

Consider observing $d_1$ binary time series over time horizon $T$, among which there exist lagged mutual-exciting effects and such effects have a finite memory depth $\tau \geq 1$. 
To be precise, consider we are given history $\{y_t^{(i)}: t = 1-\tau \dots, 0\}$ and observe $\{y_t^{(i)}: t = 1 \dots, T\}$ for $i \in \{1,\dots,d_1\}$, where $y_t^{(i)} = 1$ means type-$i$ event occurs at time $t$ and zero otherwise. In the following, we will refer to those events as node variables and our goal is to recover the mutual-excitation graph over those $d_1$ nodes.

We adopt the discrete-time Bernoulli process \citep{juditsky2020convex} and model the probability of $i$-th event's occurrence at time step $t \in \{1,\dots,T\}$ via the following generalized linear model:

\begin{align}\label{eq:GLM_summation_form}
    \mathbb{P} \left( y_t^{(i)} =1 | \mathcal{H}_{t-1} \right) = g\bigg( \nu_i +  \sum_{j = 1}^{d_1} \sum_{k = 1}^{\tau} \alpha_{ijk} y_{t-k}^{(j)} \bigg),
\end{align}
where $\mathcal{H}_t$ denotes all history observations up to time $t$. Here, $\nu_i \geq 0$ reflects the deterministic background intensity, and $\alpha_{ijk} \geq 0$ represents the magnitude of triggering effect from the $j$-th node variable to the $i$-th node variable at time lag $k$. Link function $g: \mathbb{R} \rightarrow [0,1]$ can be non-linear, such as sigmoid link function $g(x) = 1/(1+e^{-x})$ on domain $x \in \RR$ and $g(x) = 1 - e^{-x}$ on domain $x \in [0,\infty)$; also, it can be linear $g(x) = x$ on domain $x \in [0,1]$, which reduces our GLM to the simple linear model. 

For brevity, we use $w_{t - \tau : t-1}$ to denote the observations from time $t - \tau$ to $t-1$ and $\theta_i \in \RR^d$ (where $d = 1 + \tau d_1$ denotes the dimensionality) to denote the problem parameter:
\begin{align*}
    w_{t - \tau : t-1} & = \big(1,y_{t-1}^{(1)},\dots,y_{t - \tau}^{(1)},\dots,y_{t-1}^{(d_1)},\dots,y_{t - \tau}^{(d_1)}\big)^\T, \\
    \theta_i &= (\nu_i,\alpha_{i11},\dots,\alpha_{i1\tau},\dots,\alpha_{id_11},\dots,\alpha_{id_1\tau})^\T,
\end{align*}
where superscript $^\T$ denotes vector/matrix transpose.
Parameter $\theta_i$ summarizes the influence from all nodes to node $i$.
Now, we can rewrite \eqref{eq:GLM_summation_form} into the following compact form:
\begin{equation}\label{eq:model}
    \mathbb{P}\left(y_t^{(i)}=1 \Big| w_{t - \tau : t-1} \right) = g\left(w_{t - \tau : t-1}^\T \theta_i\right), \quad \theta_i \in \Theta,
\end{equation}
where $\Theta \subset \RR_+^d = [0,\infty)^d$ is the feasible region and depends on the link function. For example, when $g$ is linear function, i.e., $g(x) = x$, the feasible region is
$$\Theta = \{\theta \in \RR_+^d: 0 \leq w_{t - \tau : t-1}^\T \theta \leq 1, \ t = 1,\dots,T\}.$$

\subsection{Decoupled estimation with Variational Inequality}\label{sec:decoupled_VI_estimate}
In this section, we introduce a recently developed technique \citep{juditsky2019signal,juditsky2020convex} to estimate the parameters of the GLM by solving stochastic monotone variational inequality.
For $i \in \{1,\dots,d_1\}$. we assume the feasible region $\Theta$ is convex and compact and use the weak solution to the following variational inequality as the estimator $\hat \theta_i$ (which we will refer to as VI estimator):
\begin{equation}
   \text {find } \hat \theta_i \in \Theta:\langle  F_{ T}^{(i)}(\theta_i), \theta_i-\hat \theta_i\rangle \geq 0, \ \forall \theta_i \in \Theta,   \label{VI_1}\tag*{{VI}$[ F_{T}^{(i)}, \Theta]$}
\end{equation}
where $\langle \cdot \rangle$ represents the standard inner product in Euclidean space and $F_{T}^{(i)}(\theta_i)$ is the empirical vector field and defined as:
\begin{equation}\label{eq:empirical_vec_field}
    F_{T}^{(i)}(\theta_i) = \frac{1}{T} \sum_{t=1}^T w_{t - \tau : t-1} \left( g\left(w_{t - \tau : t-1}^\T \theta_i\right) -  y_t^{(i)} \right).
\end{equation}
As we can see, the statistical inference for each node can be {\it decoupled} and therefore we can perform the computation in parallel and simplify the analysis.

The intuition behind this method is straightforward. Let us consider the global counterpart of the above vector field, whose root is the unknown ground truth $\theta^\star_{i}$,
\begin{align*}
    F^{(i)}(\theta_i) = \mathbb{E}_{(w,y^{(i)})} \left[w \left( g\left(w^\T \theta_i\right) -  y^{(i)} \right)\right]  = \mathbb{E}_{(w,y^{(i)})} \left[w \left( g\left(w^\T \theta_i\right) -   g\left(w^\T \theta^\star_{i}\right) \right)\right].
\end{align*}
Although we cannot access this global counterpart, by solving the empirical one \ref{VI_1} we could approximate the ground truth very well. We will show how well this approximation can be by generalizing the parameter recovery guarantee in \citet{juditsky2020convex} to handle general non-linear link functions in Section~\ref{sec:theory}.

\section{Proposed Method}\label{sec:method}

As illustrated in Figure~\ref{fig:exp1_fitted_graph_illus}, it is difficult to recover the true graph structure in the presence of limited data. However, with prior knowledge on the potential graph structure, such as the directed acyclic graph structure, we can use regularization to encourage such structure and improve the recovery accuracy. Here, we will present our proposed data-adaptive linear regularization to encourage the DAG structure and show how to leverage such constraint (or rather, penalty) in the VI estimator. 

\subsection{Data-adaptive linear cycle elimination regularization}\label{sec:DAG_linear_formulation}
Consider the graphs induced by the estimated adjacency matrices $\hat A_{\ell} = (\hat \alpha_{ij \ell}) \in \RR^{d_1 \times d_1}, \ell \in \{1,\dots,\tau\}$, using estimator \ref{VI_1}. 
To return a DAG, cycles in those estimated graphs are undesirable and should be removed. 

First, let us formally define cycles: for positive integer $L \geq 2$, if there exist $\ell \in \{1,\dots,\tau\}$ and mutually different indices $i_1, \dots, i_L \in \{1,\dots,d_1\}$ such that 
$$\hat \alpha_{i_{1} i_{L} \ell} > 0, \quad \hat \alpha_{i_{k+1} i_{k} \ell} > 0, \quad k \in \{1,\dots,L-1\},$$
then we say there exists a {\it length-$L$ (directed) cycle} in the directed graphs induced by $\hat A_\ell$'s. In particular, for $L=1$ case, we say there is a {\it length-$1$ cycle (or lagged self-exciting component)} if there exist $\ell \in \{1,\dots,\tau\}$ and index $i \in \{1,\dots,d_1\}$ such that 
$\hat \alpha_{ii \ell} > 0$. 

To remove those cycles, we consider all possible length-1, 2 and 3 cycles in those estimated graphs, whose indices are denoted as follows: for all $\ell \in \{1,\dots,\tau\}$, 
\begin{align*}
    I_{1,\ell} &= \big\{i: \ \hat \alpha_{i i \ell} > 0\big\}, \quad I_{2,\ell} = \big\{(i,j): i \not= j, \ \hat \alpha_{i j \ell}, \hat \alpha_{j i \ell} > 0\big\},  \\ 
    I_{3,\ell} &= \big\{(i,j, k): i,j,k \text{ mutually different},   \hat \alpha_{i j \ell}, \hat \alpha_{j k \ell}, \hat \alpha_{k i \ell} > 0\big\}.
\end{align*}
Intuitively, in each length-2 (or 3) cycle of those estimated graphs, the edge with the least weight could be caused by noisy observation, meaning that we should remove such edge to eliminate the corresponding cycle. To do so, we impose the following {\it data-adaptive linear cycle elimination constraints}, aiming to shrink the weight of those ``least important edges'' in the cycle: for all $\ell \in \{1,\dots,\tau\}$, 
\begin{equation}\label{eq:linear_constraint}
    \begin{split}
\alpha_{i j \ell} + \alpha_{j i \ell} &\leq \delta_{2,\ell}(i,j), \quad (i,j) \in I_{2, \ell}, \\
     \alpha_{i j \ell} + \alpha_{j k \ell} + \alpha_{k i \ell} &\leq  \delta_{3,\ell}(i,j,k), \quad (i,j,k) \in I_{3, \ell}, 
    \end{split}
\end{equation}
where the {\it adaptive constraint strength parameters} are
\begin{equation*}
    \begin{split}
        \delta_{2,\ell}(i,j) &= \hat \alpha_{i j \ell} + \hat \alpha_{j i \ell} - \min\{\hat \alpha_{i j \ell}, \hat \alpha_{j i \ell}\}, \\
    \delta_{3,\ell}(i,j,k) &= \hat \alpha_{i j \ell} + \hat \alpha_{j k \ell} + \hat \alpha_{k i \ell} - \min\{\hat \alpha_{i j \ell}, \hat \alpha_{j k \ell}, \hat \alpha_{k i \ell}\}.
    \end{split}
\end{equation*}

\subsection{Joint VI estimator with penalty}\label{subsec:joint_est}

Different from the decoupled learning approach in Section~\ref{sec:decoupled_VI_estimate}, parameters $\theta_1,\dots,\theta_{d_1}$ should be estimated jointly in order to account for the desired DAG structure. We concatenate the parameter and response vectors into matrices as follows:
$$\theta = (\theta_1,\dots,\theta_{d_1}) \in \RR^{d \times d_1}, \ Y = (Y_{1:T}^{(1)},\dots,Y_{1:T}^{(d_1)}) \in \RR^{T \times d_1},$$
where $Y_{1:T}^{(i)} = (y_1^{(i)},\dots,y_T^{(i)})^\T$.
The feasible region of the concatenated parameter $\Tilde{\Theta}$ is then defined as follows:
$$\Tilde \Theta = \{\theta = (\theta_1,\dots,\theta_{d_1}): \theta_i \in \Theta, \ i = 1,\dots,d_1\}.$$

One natural idea to incorporate the data-adaptive linear constraints \eqref{eq:linear_constraint} is to add them into the feasible region $\Tilde \Theta$, which will not change the convexity of this region. However, as we typically treat the empirical vector field as the gradient filed and perform projected gradient descent (PGD) to numerically solve for the VI estimator in practice \citep{juditsky2019signal}, adding more constraints to feasible region will make the projection harder to implement; one can see a special case on how to use PGD to solve for VI estimator in Appendix~\ref{appendix:eg}. Alternatively, we propose to transfer those constraints into penalty and add the derivative of the penalty term to the vector field.
To be precise, we propose an {\it data-adaptive linear penalized VI estimator}, which is the weak solution to the following Variational Inequality:
\begin{equation*}
   \text {find } \hat \theta \in \tilde \Theta: \langle \operatorname{vec}(F_{T}^{\rm AL}(\theta)), \operatorname{vec}(\theta-\hat \theta) \rangle \geq 0, \quad \forall \theta \in \tilde \Theta, 
\end{equation*}
where $\operatorname{vec}(A)$ is the vector of columns of $A$ stacked one under the other. The data-adaptive linear penalized vector filed $F_{T}^{\rm AL}(\theta)$ is defined as follows:

\begin{align}
     F_{T}^{\rm AL} \ (\theta) =  & F_{T}(\theta) + \lambda  \sum_{\ell=1}^\tau \Bigg(  \sum_{i \in I_{1,\ell}} \frac{e_{f_{i,\ell},d} e_{i,d_1}^\T}{\hat \alpha_{ii \ell}} + \sum_{i \not \in I_{1,\ell}} \frac{e_{f_{i,\ell},d} e_{i,d_1}^\T}{\Lambda}  \label{eq:empirical_VI_whole_AL} \\
     + & \sum_{(i,j) \in I_{2, \ell}} \frac{e_{f_{j,\ell},d} e_{i,d_1}^\T + e_{f_{i,\ell},d} e_{j,d_1}^\T}{\delta_{2,\ell}(i,j)} +  \sum_{(i,j,k) \in I_{3, \ell}} 
\frac{e_{f_{j,\ell},d} e_{i,d_1}^\T + e_{f_{k,\ell},d} e_{j,d_1}^\T + e_{f_{i,\ell},d} e_{k,d_1}^\T}{\delta_{3,\ell}(i,j,k)}\Bigg), \nonumber
\end{align}
where the ``concatenated empirical vector field'' $F_T(\theta)$ is
\begin{equation}\label{eq:empirical_VI_whole}
    F_{T}(\theta) = (F_{T}^{(1)}(\theta_1),\dots,F_{T}^{(d_1)}(\theta_{d_1})) \in \RR^{d \times d_1},
\end{equation}
and the empirical vector field $F_{T}^{(i)}(\theta_i)$ is defined in \eqref{eq:empirical_vec_field}. We denote $e_{i,d} \in \RR^{d}$ to be the standard basis vector with its $i$-th element being one and define $f_{j,\ell} = 1+ (j-1)\tau + \ell$, which gives us 
$$e_{f_{j,\ell},d}^\T \theta e_{i,d_1} = \alpha_{i j \ell}, \quad \nabla_{\theta} (e_{f_{j,\ell},d}^\T \theta e_{i,d_1}) = e_{f_{j,\ell},d} e_{i,d_1}^\T.$$

\begin{remark}[Interpretation of the penalized vector filed] 
In the above penalized vector filed, the penalties at the end of the first line and the beginning of the second are very similar to adaptive Lasso \citep{zou2006adaptive}, aiming to remove all lagged self-exiting components.
Intuitively, the smaller the adaptive constraint strength parameters are, the stronger penalties should be applied, which is why those adaptive constraint strength parameters appear in the denominators. 
\end{remark}

\subsubsection{Selection of hyperparameter}
\begin{figure}[!htp]
\centering
{\includegraphics[width = .8\textwidth]{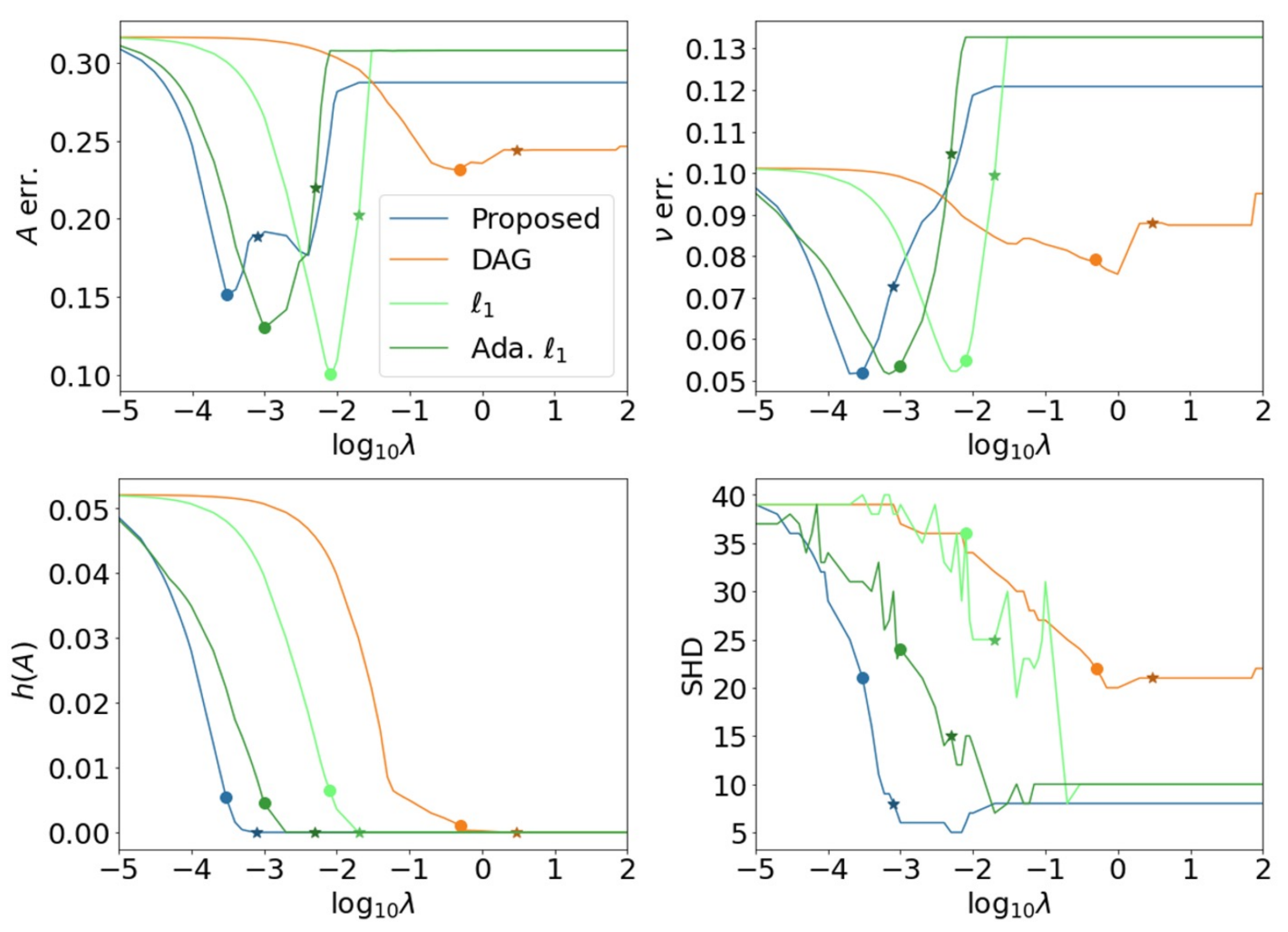}}
\caption{Illustration of the hyperparameter selection. We plot the trajectories of four performance metrics w.r.t. hyperparameter $\lambda$ for the example in Figure~\ref{fig:exp1_fitted_graph_illus}. In our numerical simulation, $\lambda$ is selected to be the smallest one which satisfies $h(A) \leq 10^{-8}$. The selected $\lambda$ is marked with a star; in addition, we mark the $\lambda$ which minimizes $A$ err. with a dot.}\label{fig:exp1_hyperparameter}
\end{figure}

Hyperparameters $\lambda$ and $\Lambda$ are tunable and control the penalty strength. In practice, hyperparameter $\Lambda$ is usually set to be a small number, such as $10^{-3}$, to ensure there will not exist self-exciting components, whereas $\lambda$ is selected based on the continuous DAG characterization \citep{zheng2018dags}. To be precise, let us consider the $\tau = 1$ special case in the illustrative example in Figure~\ref{fig:exp1_fitted_graph_illus}, and use $A = (\alpha_{ij})$ to denote $A_1 = (\alpha_{ij1})$ for brevity. The DAG characterization of a graph induced by adjacency matrix $A$ is: 

\begin{equation}\label{eq:h}
    h(A) = \operatorname{tr}(e^A) - d,
\end{equation}
where $\operatorname{tr}(e^A)$ is the trace of matrix exponential of $A$.
For $A \in \RR_+^{d_1 \times d_1}$, we have $h(A) \geq 0$ and $h(A) = 0$ if and only if the directed graph induced by adjacency matrix $A$ is a DAG. Therefore, $h(A)$ can measure the ``DAG-ness'' of $A$. We study how the performances vary with respect to (w.r.t.) hyperparameter $\lambda$ in Figure~\ref{fig:exp1_hyperparameter}; the performance evaluation metrics are: (i) matrix $F$-norm of the mutual-exciting matrix estimation error ($A$ err.), (ii) the $\ell_2$ norm of the background intensity estimation error ($\nu$ err.), (iii) ``DAG-ness'' of estimated adjacency matrix $h(A)$, and (iv) Structural Hamming Distance between the estimated adjacency matrix and the ground truth.

From Figure~\ref{fig:exp1_hyperparameter}, we can observe that the $\lambda$ which minimizes the $A$ err. (marked with a dot) typically does not give the best structural recovery (i.e., the smallest SHD); $A$ err. cannot be used to select hyperparameter anyways since its calculation requires knowledge on the ground truth.
Fortunately, we observe that the SHD converges (to its near optimal value) almost the same time when the ``DAG-ness'' measure $h(A)$ converges to zero.
Therefore, we propose to select $\lambda$ as {\it the smallest one which satisfies that $h(A) \leq \texttt{thres.}$}, where $\texttt{thres.}$ is again user-specified. Later in our numerical experiments, we will show how hyperparameter $\texttt{thres.}$  controls the balance between structural recovery (SHD) and weight recovery ($A$ err.).

\section{Theoretical Analysis}\label{sec:theory}

Here, we extend the non-asymptotic recovery guarantee of \ref{VI_1} for linear link function case in \cite{juditsky2020convex} to general non-linear link function case by imposing the following assumption:

\begin{assumption}\label{assumption:vector_field}
The link function $g(\cdot)$ is continuous and monotone, and the vector field $G(\theta) = \mathbb{E}_w[wg(w^\T\theta)]$ is well defined (and therefore monotone along with $g$). Moreover, $g$ is differentiable and has uniformly bounded first order derivative $m_g \leq |g'|\leq M_g$ for $0<m_g\leq M_g$.
\end{assumption}

The non-asymptotic upper bound on estimation error $\norm{\hat \theta_i - \theta^\star_{i}}_2$, where $\theta^\star_{i}$ is the unknown ground truth, is given by:

\begin{theorem}\label{thm:upper_err_bound}
Under Assumption~\ref{assumption:vector_field}, for $i \in \{1,\dots,d_1\}$ and any $\varepsilon \in (0,1)$, with probability at least $1-\varepsilon$,
the $\ell_2$ estimation error of \ref{VI_1} can be upper bounded as follows:
$$\norm{\hat \theta_i -  \theta^\star_{i}}_2 \leq \frac{1}{m_g \lambda_1} \sqrt{\frac{d\log (2d/\varepsilon)}{T }},$$
where $\lambda_1$ is the smallest eigenvalue of $\mathbb{W}_{1:T} = \sum_{t=1}^T w_{t - \tau : t-1}w_{t - \tau : t-1}^\T/T$.
\end{theorem}

As pointed out in \cite{juditsky2020convex}, $\mathbb{W}_{1:T} \in \mathbb{R}^{d \times d}$ will be full rank when $T$ is sufficiently large, i.e., with high probability, $\lambda_1$ will be a positive constant. The complete proof of the above theorem can be found in Appendix~\ref{appendix:proof}. One pitfall of the theoretical analysis is the lack of guarantee for the proposed data-adaptive linear regularizer and we leave this part for future discussion. In the following, we will use numerical experiments to show the good performance of our method.

\section{Numerical Experiments}\label{sec:simulation}

In this section, we provide more numerical experiments to show the effectiveness of our proposed method. We will 1) show its competitive performance under various settings and 2) study the effect regularization strength hyperparameter. In our numerical simulation, we consider $\tau = 1$ case for simplicity and choose SHD (for structural recovery) and $A$ err. (for weight recovery) as the primary performance metrics. We report the mean and standard deviation of those metrics over 200 independent trials. Complete details, such as random DAG generation, can be found in Appendix~\ref{appendix:exp}.

Let us begin with presenting benchmark methods.
The idea of transferring constraint into penalty by adding the penalty's derivative to the vector filed opens up possibilities to consider various type of DAG-inducing penalties when using the VI estimator, e.g., the continuous DAG penalty \cite{zheng2018dags} and the adaptive Lasso \cite{zou2006adaptive}. As mentioned earlier, we use $A = (\alpha_{ij})$ to denote $A_1 = (\alpha_{ij1})$ for brevity; in addition, we denote $J = (\mathbf{0}_{d_1}, I_{d_1}) \in \RR^{d_1 \times d}$ such that we have $J \theta = A^\T$.

\paragraph{\it Continuous DAG regularization.}
The DAG characterization \eqref{eq:h} has the following closed-from derivative:
$$\nabla h(A)=\left(e^{A}\right)^{\T}.$$
Inspired by \cite{ng2020role} who treated the DAG characterization directly as a penalty, we take advantage of the differentiability of the DAG penalty and add its derivative to the concatenated field $F_T(\theta)$ \eqref{eq:empirical_VI_whole}, which will later be treated as the gradient field when we use PGD to solve for the estimator. More precisely, the DAG-penalized vector field $F_{T}^{\rm DAG}(\cdot)$ is defined as follows:
\begin{equation*}
    F_{T}^{\rm DAG}(\theta) = F_{T}(\theta) \ +  \lambda J^\T \nabla h(J \theta) = F_{T}( \theta) \ +  \lambda J^\T e^{ A}.
\end{equation*}

\paragraph{\it $\ell_1$ regularization.}
We adopt the $\ell_1$ penalty as another benchmark method, which will encourage a sparse structure on the adjacency matrix $A$ and in turn eliminates cycles. To be precise, the $\ell_1$ penalized vector filed is defined as follows:
\begin{equation}\label{eq:empirical_VI_whole_l1}
    F_{T}^{\ell_1}(\theta) = F_{T}(\theta) +  \lambda J^\T \nabla (|J \theta|_1),
\end{equation}
where $|\cdot|_1$ is the summation of all entries' absolute values.

\paragraph{\it Adaptive Lasso.}
As a variant of $\ell_1$ regularization, adaptive $\ell_1$ regularization, or adaptive Lasso \cite{zou2006adaptive}, replaces $\lambda |\alpha_{ij}|$ with $\frac{\lambda}{\hat \alpha_{ij}}|\alpha_{ij}|$ in \eqref{eq:empirical_VI_whole_l1}. In addition, for $\hat \alpha_{ij} = 0$ case, we use a simple remedy by adding penalty term $\frac{\lambda}{\Lambda} |\alpha_{ij}|$ as in \eqref{eq:empirical_VI_whole_AL} to restrict $\alpha_{ij}$ to be zero.

As shown in Figure~\ref{fig:exp1_fitted_graph_illus}, our proposed data-adaptive linear approach has superior performance compared with the aforementioned DAG-inducing penalties. We will give more numerical evidence to support this in the following.

\begin{figure}[!htp]
\centering
{\includegraphics[width = .8\textwidth]{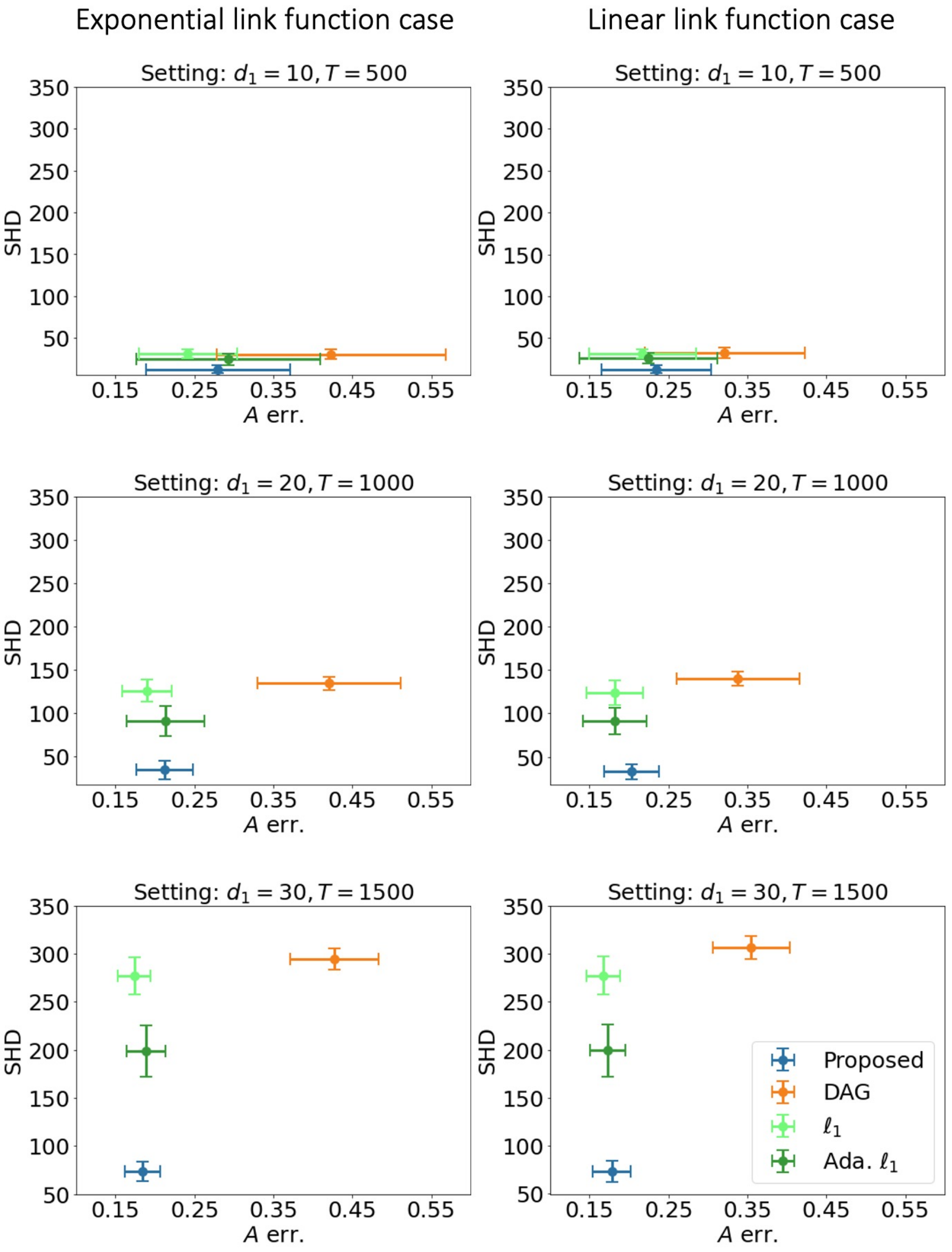}}
\caption{Comparison among different types of regularization in DAG recovery. We plot the mean (dot) and standard deviation (error bar) of matrix $F$-norm of the self- and mutual-exciting matrix estimation error ($A$ err.) and Structural Hamming Distance (SHD) over $200$ independent trials for various types of regularization. Hyperparameter $\lambda$ is selected to be the smallest one which satisfies $h(A) \leq 10^{-4}$. For each regularization, the closer it is to the origin, the better it is. We can observe that our proposed data-adaptive linear regularization performs the best (especially in higher dimensional case).}\label{fig:exp2_repeat200}
\end{figure}

\subsection{Experiment 1} First, we show the superior performance of our proposed method under settings $(d_1,T) \in \{(10,500), (20,1000), (300,1500)\}$; results are reported in Figure~\ref{fig:exp2_repeat200}. We observe that our proposed method achieves the best structural recovery among all methods, especially in higher dimensions. Besides, $\ell_1$ regularization does well in weight recovery but poorly in structural recovery. As a comparison, our proposed method achieves comparable weight recovery accuracy with $\ell_1$ regularization but much better structural recovery accuracy. On the contrary, DAG regularization is completely dominated by our proposed method, potential due to the non-convexity incurred by the DAG characterization \eqref{eq:h}; adaptive $\ell_1$ regularization achieves improved structural recovery accuracy compared with $\ell_1$ regularization, but is again dominated by our proposed method in most cases. As a sanity check, we observe the $A$ err.'s are all on the same scale for different dimension cases --- this is because we normalize each row of $A$ to sum to one to make sure it stays within the feasible region for linear link function case.

For completeness, we also report the $\nu$ err. and the ``DAG-ness'' measure $h(A)$ in Table~\ref{table:exp2_penalty_comparison} in Appendix~\ref{appendix:exp}. Those results do not only further validate our aforementioned observations, but also show $\ell_1$ regularization does the best in returning a DAG (even better than DAG regularization) but cannot return an accurate graph structure. This agrees with our illustration in Figure~\ref{fig:exp1_fitted_graph_illus} --- it does very well in encouraging sparse structure, but may shrink some important edges' weights to zeros.

\subsection{Experiment 2} 
\begin{figure}[!htp]
\centering
{\includegraphics[width = .8\textwidth]{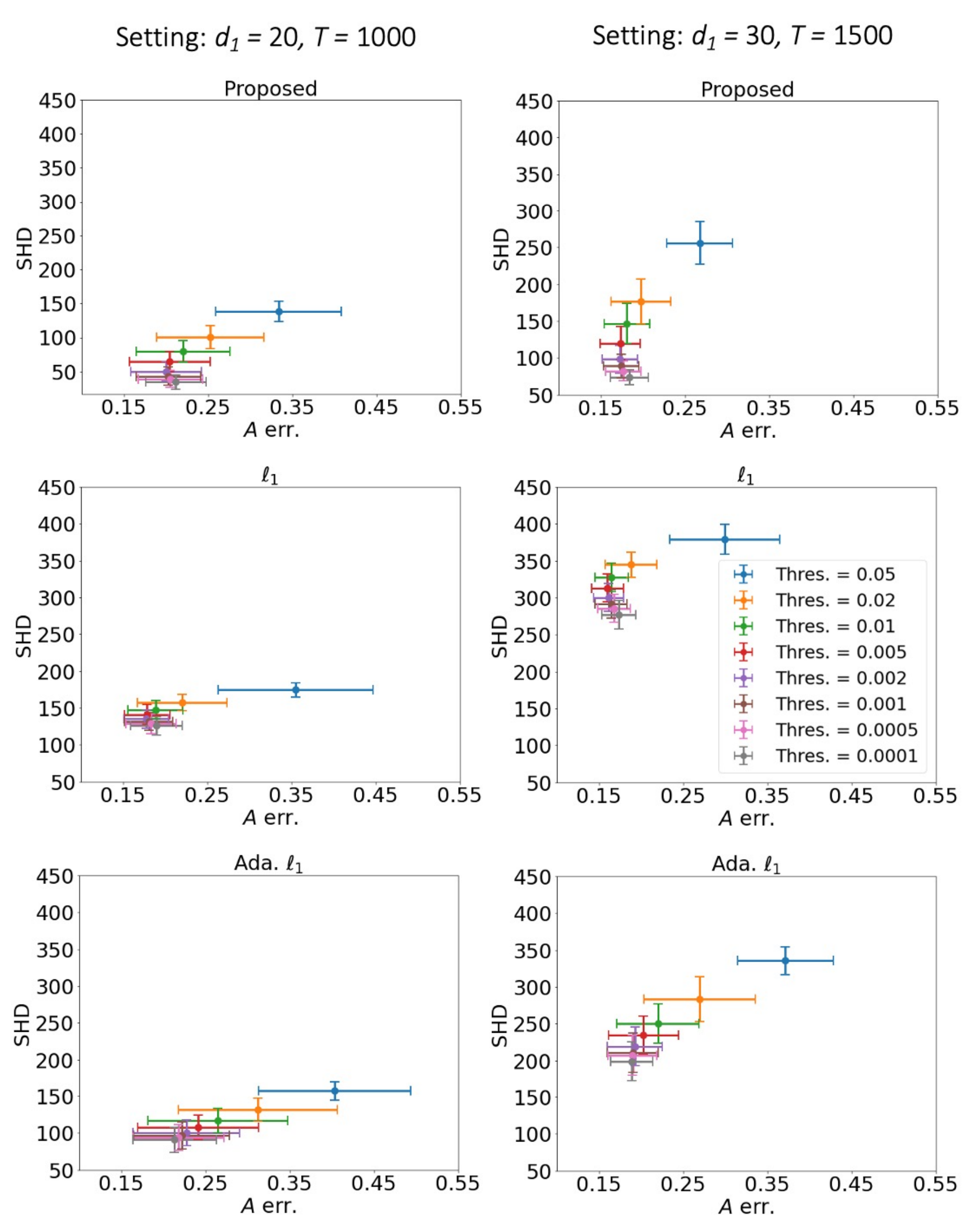}}
\caption{Effect of hyperparameter. We consider the VI estimator with exponential link function. Regularization strength hyperparameter $\lambda$ is selected to be the smallest one which satisfies that $h(A)$ is smaller than a given threshold ($\texttt{thres.}$). We plot the mean (dot) and standard deviation (error bar) of $A$ err. and SHD over $200$ independent trials for different choices of this threshold. We can observe that smaller $\texttt{thres.}$ typically leads to better SHD.}\label{fig:exp3_explink}
\end{figure}

We now study the effect of the hyperparameter $\texttt{thres.}$ introduced in Section~\ref{subsec:joint_est}. We plot the SHD and $A$ err. in Figure~\ref{fig:exp3_explink} for the exponential link function case; for completeness, we report the result for linear link function in Figure~\ref{fig:exp3_linearlink} in Appendix~\ref{appendix:exp}. From both figures, we can observe that: (i) On one hand, smaller $\texttt{thres.}$ does give better SHD. (ii) On the other hand, $A$ err. exhibits a U-shape property w.r.t. $\texttt{thres.}$, which agrees with the U-shape curves for both $A$ err. and $\nu$ err. w.r.t. $\lambda$ in Figure~\ref{fig:exp1_hyperparameter} and suggests that there could exist one optimal hyperparameter in outputting the smallest $A$ err.; however, it is an open problem on how to select it to minimize $A$ err. --- one possible approach is through the norm of empirical vector field, since it is treated as the gradient field in PGD. Nevertheless, we mainly focus on the structural recovery (i.e., SHD), and it is safe to choose a sufficient small $\texttt{thres.}$ (e.g., $\texttt{thres.} = 10^{-4}$) in practice. 

\section{Conclusion}
\label{sec:conclusion}

In this work, we go beyond the continuous but non-convex optimization approach for structural learning \cite{zheng2018dags} and formulate the DAG learning problem as a general convex optimization problem. Our theoretical analysis for the VI estimator extends the recovery guarantee in \cite{juditsky2020convex} to the general non-linear monotone ink function cases and our numerical experiments show our method's superior performance over existing methods in structural learning, opening up possibility for future work to adopt this method in a wide range of applications.

\newpage 

\bibliographystyle{plainnat}
\bibliography{ref}

\newpage 

\appendix
\section{Additional Technical Details}\label{appendix:proof}

We begin with defining an auxiliary vector field 
$$\Tilde{F}_{T}^{(i)}(\theta_i) = \frac{1}{T} \sum_{t=1}^T w_{t - \tau : t-1}( g(w_{t - \tau : t-1}^\T \theta_i) -  g(w_{t - \tau : t-1}^\T \theta^\star_{i}) ),$$
where $\theta^\star_{i}$ is the unknown ground truth.
This vector field has a nice property that its unique root/weak solution to corresponding VI is $\theta^\star_{i}$, whereas the VI estimator $\hat \theta_i$ is the root of ${F}_{T}^{(i)}(\theta_i)$.
Next, we bound the difference between $\hat \theta_i$ and $\theta^\star_{i}$ by bounding the difference between the empirical vector field ${F}_{T}^{(i)}(\theta_i)$ and the auxiliary vector field $\Tilde{F}_{T}^{(i)}(\theta_i)$, i.e.,
$$\Delta^{(i)} = {F}_{T}^{(i)}(\theta_i) - \Tilde{F}_{T}^{(i)}(\theta_i) = {F}_{T}^{(i)}(\theta^\star_{i}).$$
\begin{proposition}\label{lma:bound_delta}
Under Assumption~\ref{assumption:vector_field}, for any $\varepsilon \in (0,1)$, with probability at least $1-\varepsilon$, for $i \in \{1,\dots,d_1\}$, $\Delta^{(i)}$ can be bounded as follows:
\begin{equation}\label{eq:bound_inf}
    \norm{\Delta^{(i)}}_\infty \leq  \sqrt{{\log(2d/\varepsilon)}/{T}}.
\end{equation}
Moreover, this implies
\begin{equation}\label{eq:bound_l2}
    \norm{\Delta^{(i)}}_2 \leq  \sqrt{N{\log(2d/\varepsilon)}/{T}}.
\end{equation}
\end{proposition}

\begin{proof}
Denote random vector 
$$\xi_t= w_{t - \tau : t-1} \left( g\left(w_{t - \tau : t-1}^\T \theta^\star_{i}\right) -  y_t^{(i)} \right).$$ 
We can re-write $\Delta^{(i)} = \sum_{t=1}^T \xi_t/T.$ Define $\sigma$-field $\mathcal{F}_t = \sigma (W_t)$, and $\mathcal{F}_0 \subset \mathcal{F}_1 \subset \cdots \mathcal{F}_T$ form a filtration. We can show
\begin{align*}
    \mathbb{E}[\xi_t|\mathcal{F}_{t-1}] &= 0, \\
    {\rm Var}(\xi_t|\mathcal{F}_{t-1}) &=  g\left(w_{t - \tau : t-1}^\T \theta_i\right) \left(1-g\left(w_{t - \tau : t-1}^\T \theta_i\right)\right) \leq 1/4.
\end{align*}
This means $\xi_t, t \in \{1,\dots,T\},$ is a Martingale Difference Sequence. Moreover, its infinity norm is upper bounded by one almost surly since it only consists of binary elements. Therefore, Azuma's inequality gives us:
$$\mathbb{P}\left( |\Delta^{(i)}_k| > u \right) \leq 2 \exp\left\{ - \frac{Tu^2}{2}\right\}, \ k = 1,\dots,d, \ \forall \  u>0,$$
where $\Delta^{(i)}_k$ is the $k$-th entry of vector $\Delta^{(i)}$. By union bound, 
$$\mathbb{P}\left( |\Delta^{(i)}_k| > u, \ k = 1,\dots,d \right) \leq 2d \exp\left\{ - \frac{Tu^2}{2}\right\}, \ \forall \  u>0.$$
By setting the RHS of above inequality to $\varepsilon$ and solving for $u$, we prove \eqref{eq:bound_inf}. Besides,
since $\norm{\Delta}_2 \leq \sqrt{d}  \norm{\Delta}_\infty$ holds for any vector $\Delta \in \mathbb{R}^d$, we can easily prove \eqref{eq:bound_l2} using \eqref{eq:bound_inf}.
\end{proof}

The proof of Proposition~\ref{lma:bound_delta} leverages the concentration property of martingales. 
By this proposition, we can now prove the non-asymptotic estimation error bound as follows:

\begin{proof}[Proof of Theorem~\ref{thm:upper_err_bound}]
Under Assumption~\ref{assumption:vector_field}, the vector field $F_{T}^{(i)}(\theta_i)$ is monotone modulus $m_g \lambda_1$, where $\lambda_1$ is the smallest eigenvalue of $\mathbb{W}_{1:T} = \sum_{t=1}^T w_{t - \tau : t-1}w_{t - \tau : t-1}^\T/T$. This can be proved as follows:
\begin{align*}
    & \left(F_{T}^{(i)}(\theta) - F_{T}^{(i)}(\theta')\right)^\T (\theta - \theta') \\
    &= \frac{1}{T} \sum_{t=1}^T  w_{t - \tau : t-1}^\T \left(\theta - \theta'\right) \left(g\left(w_{t - \tau : t-1}^\T\theta\right) - g\left(w_{t - \tau : t-1}^\T\theta'\right)\right)\\
    &\geq m_g \frac{1}{T} \sum_{t=1}^T \norm{ w_{t - \tau : t-1}^\T (\theta - \theta')}_2^2\\
    &= m_g (\theta - \theta')^\T \frac{1}{T} \sum_{t=1}^T w_{t - \tau : t-1}w_{t - \tau : t-1}^\T (\theta - \theta')\\
    & \geq m_g \lambda_1 \norm{\theta - \theta'}_2^2.
\end{align*}

In particular, we have:
$$\left({F}_{T}^{(i)}(\hat \theta_i)- {F}_{T}^{(i)}(\theta^\star_{i})\right)^\T (\hat \theta_i - \theta^\star_{i}) \geq m_g\lambda_1 \norm{\hat \theta_i -  \theta^\star_{i}}_2^2.$$
Notice that our weak solution $\hat \theta_i$ is also a strong solution to the VI since the empirical vector field is continuous, which gives us
$$\left({F}_{T}^{(i)}(\hat \theta_i)\right)^\T (\hat \theta_i - \theta^\star_{i}) \leq 0.$$
By triangle inequality, we also have $$- \left({F}_{T}^{(i)}(\theta^\star_{i})\right)^\T (\hat \theta_i - \theta^\star_{i}) = - \Delta_i^\T (\hat \theta_i - \theta^\star_{i}) \leq \norm{\Delta_i}_2 \norm{\hat \theta_i -  \theta^\star_{i}}_2.$$ 
Together with \eqref{eq:bound_l2} in Proposition~\ref{lma:bound_delta}, we complete the proof.

\end{proof}

\paragraph{\it Identifiablility of our proposed estimator.}
In addition, we can show the uniqueness, or rather, the identifiablility of the VI estimator \ref{VI_1}, which comes from the nice property of the underlying vector field. To be precise, in the proof of the above theorem, we have shown the vector field $F_{T}^{(i)}(\theta_i)$ is monotone modulus $m_g \lambda_1$ under Assumption~\ref{assumption:vector_field}. Then, the following lemma tells us that our proposed estimator is unique:

\begin{lemma}[Lemma 3.1 \cite{juditsky2019signal}]
Let $\Theta$ be a convex compact set and $H$ be a monotone vector field on $\Theta$ with monotonicity modulus $\kappa>0$, i.e.,
$$\forall \ z, z' \in \Theta, [H(z) - H(z')]^\T(z-z') \geq \kappa \norm{z-z'}_2^2.$$
Then, the weak solution $\Bar{z}$ to VI$[H,\Theta]$ exists and is unique. It satisfies:
$$H(z)^\T(z-\Bar{z}) \geq \kappa \norm{z-\Bar{z}}_2^2.$$
\end{lemma}

\section{A Special Example}\label{appendix:eg} 
\subsection{Decoupled estimation}
We consider a special case where $g(x) = x$. To make sure the model \eqref{eq:model} indeed returns a meaningful probability, we require the parameter $\theta_i$ to take value in 
$\Theta = \{\theta_i \in \RR_+^d : \ 0 \leq w_{t - \tau : t-1}^\T \theta_i \leq 1, \ t = 1,\dots,T\}.$
In this special case, the empirical vector field \eqref{eq:empirical_vec_field} becomes
\begin{equation*}\label{eq:liner}
\begin{split}
        F_{T}^{(i)}(\theta_i) & = \frac{1}{T} \sum_{t=1}^T w_{t - \tau : t-1} w_{t - \tau : t-1}^\T \theta_i - \frac{1}{T} \sum_{t=1}^T w_{t - \tau : t-1} y_t^{(i)}\\
        & =\mathbb{W}_{1:T}\theta_i - \frac{1}{T} \sum_{t=1}^T w_{t - \tau : t-1} y_t^{(i)},
\end{split}
\end{equation*}
where we denote 
\begin{equation}\label{eq:vec_field_linear}
\begin{split}
      \mathbf{w}_{1:T} &= (w_{1-\tau : 0},\dots,w_{T - \tau : T-1}) \in \mathbb{R}^{d \times T}, \\  \mathbb{W}_{1:T} &= \frac{1}{T} \mathbf{w}_{1:T} \mathbf{w}_{1:T}^\T =\frac{1}{T} \sum_{t=1}^T w_{t - \tau : t-1}w_{t - \tau : t-1}^\T \in \mathbb{R}^{d \times d}.
\end{split}
\end{equation} 
Most importantly, this vector field is indeed the gradient field of the least square objective, 
meaning that the weak solution to the corresponding VI is the following LS estimator \cite{juditsky2020convex}:
\begin{equation}\label{eq:objective:convex}
\begin{array}{rl}
\underset{\theta_i}{\mbox{min}} & \frac{1}{2T} \norm{\mathbf{w}_{1:T}^\T \theta_i - Y_{1:T}^{(i)}}_2^2, \\
\mbox{subject to} &  \theta_i \geq \mathbf{0}_T, \ \mathbf{1}_T - \mathbf{w}_{1:T}^\T \theta_i \geq \mathbf{0}_T, 
\end{array}
\end{equation}
where $Y_{1:T}^{(i)} = (y_1^{(i)},\dots,y_T^{(i)})^\T$, $\mathbf{0}_T$ and $\mathbf{1}_T$ are the column vectors of all zeros and ones in $\RR^T$, respectively, and $\norm{\cdot}_p$ denotes the vector $\ell_p$ norm.

Note that the equivalence between our proposed estimatorand LS estimator will only hold for linear link function, since the gradient field of LS objective with general link function is:
$$ \frac{1}{T} \sum_{t=1}^T w_{t - \tau : t-1} g'\left(w_{t - \tau : t-1}^\T \theta_i\right) \left( g\left(w_{t - \tau : t-1}^\T \theta_i\right) -  y_t^{(i)} \right).$$

One approach to solve \eqref{eq:objective:convex} is to leverage the well-developed optimization tools, such as \texttt{Mosek} \cite{mosek}. An alternative approach is through projected gradient descent, where the empirical vector field \eqref{eq:empirical_vec_field} is treated as the gradient. To be precise, we introduce dual variables $\eta_1 = (\eta_{1,1},\dots,\eta_{1,T})^\T$, $\eta_2 = (\eta_{2,1},\dots,\eta_{2,d})^\T$ and the Lagrangian is as follows:
\begin{equation*}
    \begin{split}
        L(\theta_i, \eta_1, \eta_2) = \frac{1}{2T} & \norm{\mathbf{w}_{1:T}^\T \theta_i - Y_{1:T}^{(i)}}_2^2 \\
        & + \eta_1^\T (\mathbf{w}_{1:T}^\T \theta_i - \mathbf{1}_T) - \eta_2^\T \theta_i.
    \end{split}
\end{equation*}
The Lagrangian dual function is $\min_{\theta_i} L(\theta_i, \eta_1, \eta_2)$. As we can see, the Lagrangian above is convex w.r.t. $\theta_i$. By setting the derivative of $L(\theta_i, \eta_1, \eta_2)$ w.r.t. $\theta_i$ to zero, we have
$$
\hat \theta_i = \frac{1}{T} \mathbb{W}_{1:T}^{-1} \left(\mathbf{w}_{1:T} Y_{1:T}^{(i)}/T - \eta_1 \right) + \eta_2,
$$
which minimizes the Lagrangian dual function. As pointed out in \cite{juditsky2020convex}, $\mathbb{W}_{1:T} \in \mathbb{R}^{d \times d}$ will be full rank with high probability when $T$ is sufficiently large, and therefore $\mathbb{W}_{1:T}^{-1}$ exists.
By plugging $\hat \theta_i$ into the Lagrangian dual function $\min_{\theta_i} L(\theta_i, \eta_1, \eta_2)$, we give the dual problem as follows:
\begin{equation*}
\begin{array}{rl}
\underset{\eta_1, \eta_2}{\mbox{max}} & L(\hat \theta_i, \eta_1, \eta_2), \\
\mbox{subject to} & \eta_1, \eta_2 \geq \mathbf{0}_T.
\end{array}
\end{equation*}
As we can see, this dual problem can be easily solved by PGD. 

\subsection{Joint estimation}
Now let us consider the joint estimation, where the vector field $F_T(\theta)$ \eqref{eq:empirical_VI_whole} can be expressed as follows:
$$F_{T}(\theta) = \frac{1}{T} \mathbf{w}_{1:T} \mathbf{w}_{1:T}^\T \theta - \frac{1}{T} \mathbf{w}_{1:T} Y = \mathbb{W}_{1:T}\theta - \frac{1}{T} \mathbf{w}_{1:T} Y,$$
where $\mathbf{w}_{1:T} \in \RR^{d \times T}$ is defined in \ref{eq:vec_field_linear}. Similar to the example in decoupled estimation, the above vector field is the gradient field of the least square objective, and our proposed estimator boils down to LS estimator, which solves the following penalized optimization problem:
\begin{equation}\label{eq:LS_DAG_penalized_form}
\begin{split}
\min_{\theta \in \tilde \Theta}
      & \frac{1}{2T} \norm{\mathbf{w}_{1:T}^\T \theta - Y}_F^2  +  \lambda \sum_{\ell=1}^\tau \Bigg(  \sum_{i \in I_{1,\ell}} \frac{e_{f_{i,\ell},d}^\T \theta e_{i,d_1}}{\hat \alpha_{ii \ell}}  + \sum_{i \not \in I_{1,\ell}} \frac{e_{f_{i,\ell},d}^\T \theta e_{i,d_1}}{\Lambda} \\
    + & \sum_{(i,j) \in I_{2, \ell}} \frac{e_{f_{j,\ell},d} ^\T \theta e_{i,d_1} + e_{f_{i,\ell},d}^\T \theta e_{j,d_1}}{\delta_{2,\ell}(i,j)}  +  \sum_{(i,j,k) \in I_{3, \ell}} 
\frac{e_{f_{j,\ell},d}^\T \theta e_{i,d_1} + e_{f_{k,\ell},d}^\T \theta e_{j,d_1} + e_{f_{i,\ell},d}^\T \theta e_{k,d_1}}{\delta_{3,\ell}(i,j,k)}\Bigg), 
\end{split}
\end{equation}
where $\norm{\cdot}_F$ is the matrix $F$-norm.
Therefore, \eqref{eq:LS_DAG_penalized_form} can be solved efficiently using PGD, where at each iteration the update rule is as follows:
\begin{equation*}
\begin{split}
    \hat \theta \leftarrow \hat \theta - \eta F_{T}^{\rm AL}(\hat \theta),
\end{split}
\end{equation*}
where $\eta$ is the step size/learning rate hyperparameter and $F_{T}^{\rm AL}(\cdot)$ is the penalized empirical field \eqref{eq:empirical_VI_whole_AL}. Since the prediction of the $i$-th event at time $t$ is determined by the estimated probability $w_{t - \tau:t-1}^\T \theta_i$ and a cut-off/threshold selected using the validation dataset, we can further relax the constraint $\mathbf{w}_{1:T}^\T \theta_i \leq \mathbf{1}_T$ and treat it as ``score'' instead of probability. Therefore, after the above update in each iteration, the projection onto the (relaxed) feasible region can be simply done by replacing all negative entries in $\hat \theta$ with zeros.

\section{Additional Numerical Experiments}\label{appendix:exp}

\subsection{Additional experimental details}
We first randomly generate $\nu$ and $A$ using standard uniform distribution. Next, to make sure $A$ stays in the feasible region $\Theta$ for the linear function case, we normalize each row to make sure it sums up to one. To be precise, we just update each entry in the row by dividing it with the row summation. To make sure $A$ is DAG, we (i) first ``sparse-ify'' it by setting all entries smaller than the $95\%$ percentile to zeros and (ii) next minimize the DAG characterization $h(A)$ using vanilla gradient descent (learning rate is $0.5$ and we consider in total $5000$ iterations). The reason of applying (i) is the highly non-convex optimization in (ii) --- if we do not input a highly sparse graph, then we cannot shrink the DAG characterization to exactly zero with high probability.
As for PGD approach to solve for VI estimator, we use $5 \times 10^{-3}$ as the initial learning rate and decrease it by half every $2000$ iterations (in total there are $6000$ iterations).

\subsection{Additional experimental results} 
\begin{figure*}[!htp]
\centering
{\includegraphics[width = \textwidth]{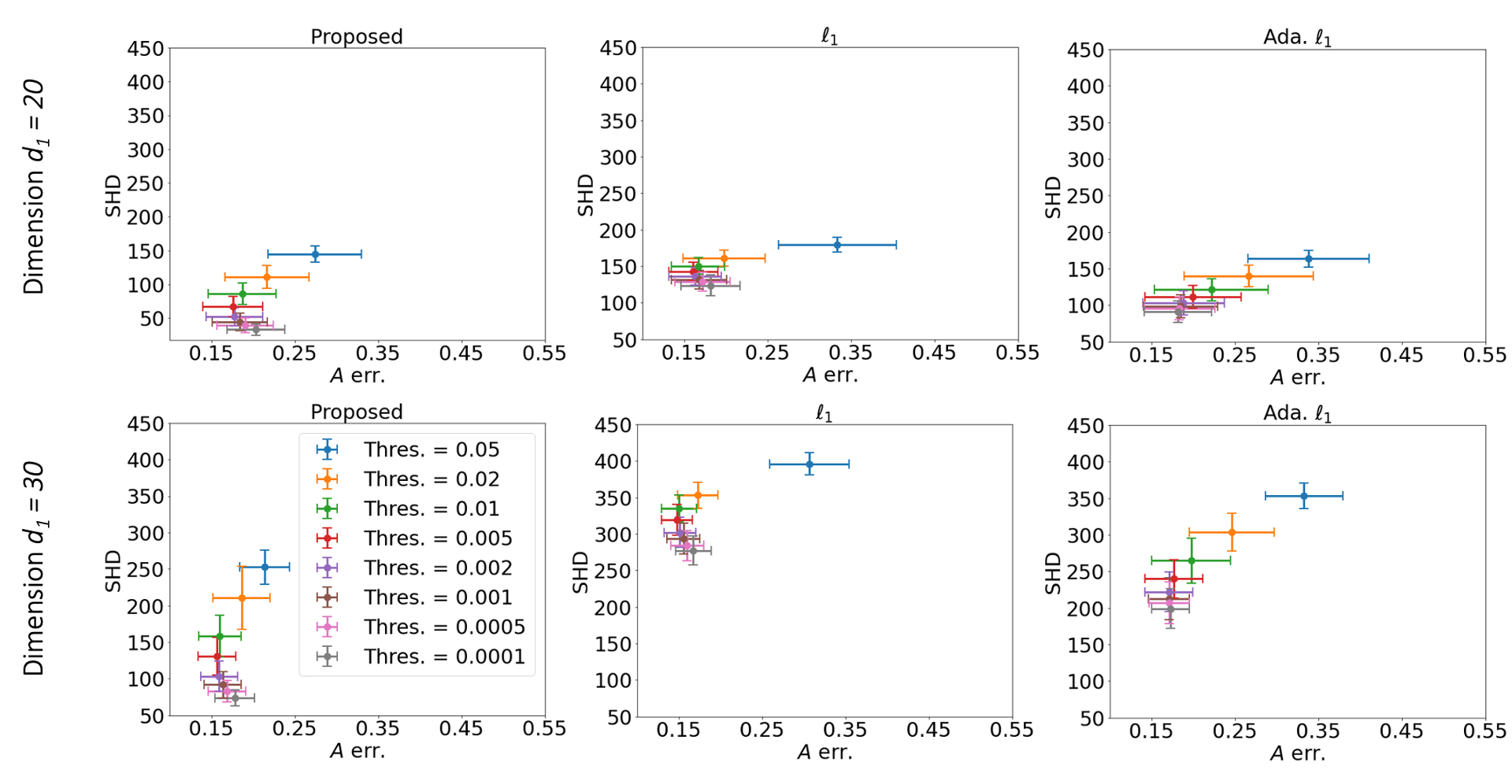}}
\caption{Effect of hyperparameter (continued). We consider the VI estimator with linear link function for completeness in this figure. We can observe similar patterns with Figure~\ref{fig:exp3_explink}.}\label{fig:exp3_linearlink}
\end{figure*}

Due to space consideration, we report the results that do not give new insights and only serve for completeness purpose here. In particular, we report all four aforementioned metrics for Experiment 1 in Table~\ref{table:exp2_penalty_comparison}, and plot the results for linear link function for Experiment 2 in Figure~\ref{fig:exp3_linearlink}; please see the interpretation of those results in Section~\ref{sec:simulation}.

\begin{table*}[htp]
\caption{Comparison of the mean (and standard deviation) of various performance metrics over $200$ trials for different types of regularization. We report the matrix $F$-norm of the self- and mutual-exciting matrix estimation error ($A$ err.), the $\ell_2$ norm of the background intensity estimation error ($\nu$ err.), the ``DAG-ness'' measured by $h(A)$ and the Structural Hamming Distance (SHD). 
}\label{table:exp2_penalty_comparison}

\begin{center}
\begin{small}
\begin{sc}
\resizebox{.8\textwidth}{!}{%
\begin{tabular}{lccccc}
\multicolumn{6}{c}{\large{Linear Link.}} \\ 
\multicolumn{6}{c}{\small{Dimension $d_1 = 10$, Time Horizon $T = 500$.}} \\ 
\toprule[1pt]\midrule[.3pt] 
Regularization & None  & Proposed & DAG & $\ell_1$ & Ada. $\ell_1$ \\
\cmidrule(l){2-6} 
\texttt{$A$ err.} & $0.3379_{(0.0988)}$ & $0.2347_{(0.0698)}$ & $0.3214_{(0.1009)}$ & $0.2172_{(0.0674)}$ & $0.2246_{(0.0872)}$ \\
\texttt{$\nu$ err.} & $0.0970_{(0.0307)}$ & $0.0661_{(0.0213)}$ & $0.0822_{(0.0266)}$ & $0.0636_{(0.0208)}$ & $0.0584_{(0.0170)}$ \\
$h(A)$ & $0.0311_{(0.0163)}$ & $0.0002_{(0.0011)}$ & $0.0053_{(0.0041)}$ & $0.0000_{(0.0000)}$ & $0.0000_{(0.0000)}$ \\
SHD & $44.3_{(5.24)}$ & $12.74_{(4.73)}$ & $32.7_{(6.34)}$ & $31.77_{(5.48)}$ & $26.22_{(6.32)}$ \\
\midrule[.3pt]\bottomrule[1pt]
\end{tabular}
}
\end{sc}
\end{small}
\end{center}

\vspace{-.3in}

\begin{center}
\begin{small}
\begin{sc}
\resizebox{.8\textwidth}{!}{%
\begin{tabular}{lccccc}
\multicolumn{6}{c}{\small{Dimension $d_1 = 20$, Time Horizon $T = 1000$.}} \\ 
\toprule[1pt]\midrule[.3pt] 
Regularization & None  & Proposed & DAG & $\ell_1$ & Ada. $\ell_1$ \\
\cmidrule(l){2-6} 
\texttt{$A$ err.} & $0.3764_{(0.0737)}$ & $0.2035_{(0.0348)}$ & $0.3382_{(0.0783)}$ & $0.1820_{(0.0357)}$ & $0.1819_{(0.0401)}$ \\
\texttt{$\nu$ err.} & $0.1729_{(0.0329)}$ & $0.0969_{(0.0253)}$ & $0.1403_{(0.0259)}$ & $0.0834_{(0.0225)}$ & $0.0735_{(0.0170)}$ \\
$h(A)$ & $0.0573_{(0.0132)}$ & $0.0000_{(0.0000)}$ & $0.0086_{(0.0012)}$ & $0.0000_{(0.0000)}$ & $0.0000_{(0.0000)}$ \\
SHD & $183.28_{(10.70)}$ & $32.99_{(8.65)}$ & $139.79_{(8.19)}$ & $123.64_{(14.34)}$ & $91.24_{(15.07)}$ \\
\midrule[.3pt]\bottomrule[1pt]
\end{tabular}
}
\end{sc}
\end{small}
\end{center}

\vspace{-.3in}

\begin{center}
\begin{small}
\begin{sc}
\resizebox{.8\textwidth}{!}{%
\begin{tabular}{lccccc}
\multicolumn{6}{c}{\small{Dimension $d_1 = 30$, Time Horizon $T = 1500$.}} \\ 
\toprule[1pt]\midrule[.3pt] 
Regularization & None  & Proposed & DAG & $\ell_1$ & Ada. $\ell_1$ \\
\cmidrule(l){2-6} 
\texttt{$A$ err.} & $0.4116_{(0.0448)}$ & $0.1782_{(0.0239)}$ & $0.3549_{(0.0491)}$ & $0.1676_{(0.0213)}$ & $0.1727_{(0.0226)}$ \\
\texttt{$\nu$ err.} & $0.2486_{(0.0334)}$ & $0.1104_{(0.0210)}$ & $0.1995_{(0.0262)}$ & $0.1013_{(0.0187)}$ & $0.0987_{(0.0190)}$ \\
$h(A)$ & $0.0774_{(0.0113)}$ & $0.0000_{(0.0000)}$ & $0.0089_{(0.0011)}$ & $0.0000_{(0.0000)}$ & $0.0000_{(0.0000)}$ \\
SHD & $411.81_{(12.18)}$ & $73.43_{(11.03)}$ & $306.52_{(11.69)}$ & $277.25_{(19.99)}$ & $199.15_{(26.89)}$ \\
\midrule[.3pt]\bottomrule[1pt]
\end{tabular}
}
\end{sc}
\end{small}
\end{center}

\vspace{0.1in}

\begin{center}
\begin{small}
\begin{sc}
\resizebox{.8\textwidth}{!}{%
\begin{tabular}{lccccc}
\multicolumn{6}{c}{\large{Exponential Link.}} \\ 
\multicolumn{6}{c}{\small{Dimension $d_1 = 10$, Time Horizon $T = 500$.}} \\ 
\toprule[1pt]\midrule[.3pt] 
Regularization & None  & Proposed & DAG & $\ell_1$ & Ada. $\ell_1$ \\
\cmidrule(l){2-6} 
\texttt{$A$ err.} & $0.4495_{(0.1457)}$ & $0.2797_{(0.0914)}$ & $0.4233_{(0.1452)}$ & $0.2417_{(0.0620)}$ & $0.2925_{(0.1167)}$ \\
\texttt{$\nu$ err.} & $0.1061_{(0.0336)}$ & $0.0720_{(0.0225)}$ & $0.0889_{(0.0285)}$ & $0.0666_{(0.0208)}$ & $0.0644_{(0.0196)}$ \\
$h(A)$ & $0.0439_{(0.0243)}$ & $0.0001_{(0.0006)}$ & $0.0046_{(0.0039)}$ & $0.0000_{(0.0000)}$ & $0.0000_{(0.0000)}$ \\
SHD & $43.75_{(5.00)}$ & $12.96_{(5.11)}$ & $30.77_{(5.71)}$ & $31.66_{(5.28)}$ & $24.7_{(6.44)}$ \\
\midrule[.3pt]\bottomrule[1pt]
\end{tabular}
}
\end{sc}
\end{small}
\end{center}

\vspace{-.3in}

\begin{center}
\begin{small}
\begin{sc}
\resizebox{.8\textwidth}{!}{%
\begin{tabular}{lccccc}
\multicolumn{6}{c}{\small{Dimension $d_1 = 20$, Time Horizon $T = 1000$.}} \\ 
\toprule[1pt]\midrule[.3pt] 
Regularization & None  & Proposed & DAG & $\ell_1$ & Ada. $\ell_1$ \\
\cmidrule(l){2-6} 
\texttt{$A$ err.} & $0.4731_{(0.0844)}$ & $0.2118_{(0.0360)}$ & $0.4206_{(0.0911)}$ & $0.1898_{(0.0310)}$ & $0.2136_{(0.0496)}$ \\
\texttt{$\nu$ err.} & $0.1897_{(0.0385)}$ & $0.0948_{(0.0201)}$ & $0.1511_{(0.0298)}$ & $0.0840_{(0.0187)}$ & $0.0813_{(0.0174)}$ \\
$h(A)$ & $0.0799_{(0.0191)}$ & $0.0002_{(0.0011)}$ & $0.0087_{(0.001)}$ & $0.0000_{(0.0000)}$ & $0.0000_{(0.0000)}$ \\
SHD & $183.83_{(10.18)}$ & $34.59_{(10.79)}$ & $134.53_{(7.80)}$ & $125.7_{(12.75)}$ & $90.8_{(17.29)}$ \\
\midrule[.3pt]\bottomrule[1pt]
\end{tabular}
}
\end{sc}
\end{small}
\end{center}

\vspace{-.3in}

\begin{center}
\begin{small}
\begin{sc}
\resizebox{.8\textwidth}{!}{%
\begin{tabular}{lccccc}
\multicolumn{6}{c}{\small{Dimension $d_1 = 30$, Time Horizon $T = 1500$.}} \\ 
\toprule[1pt]\midrule[.3pt] 
Regularization & None  & Proposed & DAG & $\ell_1$ & Ada. $\ell_1$ \\
\cmidrule(l){2-6} 
\texttt{$A$ err.} & $0.5048_{(0.0507)}$ & $0.1841_{(0.0225)}$ & $0.4277_{(0.0559)}$ & $0.1738_{(0.0205)}$ & $0.1888_{(0.0250)}$ \\
\texttt{$\nu$ err.} & $0.2743_{(0.0361)}$ & $0.1090_{(0.0170)}$ & $0.2148_{(0.0274)}$ & $0.1022_{(0.0168)}$ & $0.1032_{(0.0162)}$ \\
$h(A)$ & $0.1090_{(0.0151)}$ & $0.0000_{(0.0000)}$ & $0.0089_{(0.0010)}$ & $0.0000_{(0.0000)}$ & $0.0000_{(0.0000)}$ \\
SHD & $414.17_{(13.44)}$ & $73.75_{(10.52)}$ & $294.5_{(11.44)}$ & $277.14_{(19.12)}$ & $198.52_{(26.61)}$ \\
\midrule[.3pt]\bottomrule[1pt]
\end{tabular}
}
\end{sc}
\end{small}
\end{center}
\end{table*}

\end{document}